\documentclass[journal,transmag]{IEEEtran}

\usepackage{amsmath,amssymb,amsfonts}
\usepackage{algorithmic}
\usepackage{graphicx}
\usepackage{textcomp}
\usepackage{multirow}
\usepackage{algorithmic}
\usepackage{graphicx}
\usepackage{textcomp}
\usepackage{xcolor}
\usepackage{psfrag}
\usepackage{epsfig}
\usepackage{cite}
\usepackage{graphics}
\usepackage{graphicx}
\usepackage{color}
\usepackage{subfigure}
\usepackage{multirow}
\usepackage{booktabs}
\usepackage{bbding}
\usepackage{pifont}
\usepackage{graphicx}
\usepackage{epstopdf}
\usepackage{mmstyles}
\usepackage[ruled,vlined]{algorithm2e}
\usepackage{bm}

\newcommand{\cxmark}{\ding{55}}

\def \eg {\emph{e.g.}}

\def \etal {\emph{et al. }}

\ifCLASSINFOpdf
\else
\fi

\begin{document}
%

\title{RegFreeNet: A Registration-Free Network for CBCT-based 3D Dental Implant Planning}


\author{
\IEEEauthorblockN{Xinquan Yang\IEEEauthorrefmark{1,2},
Xuguang Li\IEEEauthorrefmark{3},
Mianjie Zheng\IEEEauthorrefmark{1},
Xuefen Liu\IEEEauthorrefmark{1},
Kun Tang\IEEEauthorrefmark{1},
Kian Ming Lim\IEEEauthorrefmark{2},\\
He Meng\IEEEauthorrefmark{3},
Jianfeng Ren\IEEEauthorrefmark{2},~\IEEEmembership{Senior Member,~IEEE}, and
Linlin Shen\IEEEauthorrefmark{1,2},~\IEEEmembership{Senior Member,~IEEE}}

\IEEEauthorblockA{\IEEEauthorrefmark{1}School of Artificial Intelligence, Shenzhen University, Shenzhen, China}
\IEEEauthorblockA{\IEEEauthorrefmark{2}School of Computer Science, University of Nottingham Ningbo China, Ningbo, China}
\IEEEauthorblockA{\IEEEauthorrefmark{3}Department of Stomatology, Shenzhen University General Hospital, Shenzhen, China}
}

\markboth{Journal of \LaTeX\ Class Files,~Vol.~14, No.~8, August~2015}%
{Shell \MakeLowercase{\textit{et al.}}: Bare Demo of IEEEtran.cls for IEEE Transactions on Magnetics Journals}

\IEEEtitleabstractindextext{%
\begin{abstract}
As the commercial surgical guide design software usually does not support the export of implant position for pre-implantation data, existing methods have to scan the post-implantation data and map the implant to pre-implantation space to get the label of implant position for training. 
Such a process is time-consuming and heavily relies on the accuracy of registration algorithm.
Moreover, not all hospitals have paired CBCT data, limitting the construction of multi-center dataset. 
Inspired by the way dentists determine the implant position based on the neighboring tooth texture, we found that even if the implant area is masked, it will not affect the determination of the implant position.
Therefore, we propose to mask the implants in the post-implantation data so that any CBCT containing the implants can be used as training data. 
This paradigm enables us to discard the registration process and makes it possible to construct a large-scale multi-center implant dataset.
On this basis, we proposes ImplantFairy, a comprehensive, publicly accessible dental implant dataset with voxel-level 3D annotations of 1622 CBCT data. 
Furthermore, according to the area variation characteristics of the tooth's spatial structure and the slope information of the implant, we designed a slope-aware implant position prediction network.
Specifically, a neighboring distance perception (NDP) module is designed to adaptively extract tooth area variation features, and an implant slope prediction branch assists the network in learning more robust features through additional implant supervision information.
Extensive experiments conducted on ImplantFairy and two public dataset demonstrate that the proposed RegFreeNet achieves the state-of-the-art performance.
\end{abstract}

}

\maketitle

\IEEEdisplaynontitleabstractindextext
\IEEEpeerreviewmaketitle

\section{Introduction}

Tooth loss and dental fractures are widespread oral health issues globally~\cite{elani2018trends,nazir2020global}, and dental implantation serves as a key restorative solution. Surgical guides are often utilized to ensure implants are placed precisely as planned, while current guide design relies on manual and operator‑dependent CBCT analysis~\cite{d2012accuracy}. Deep learning offers a promising avenue to transform this workflow by automating implant planning, determining the optimal three‑dimensional placement (including location, angle, and depth) of a dental or orthopedic implant within the patient's anatomy. Such approaches can improve the accuracy and efficiency of guided surgery~\cite{liu2021transfer,yang2023two}, while enabling patient-specific guide design, virtual preoperative planning, and simulation-based training. Integration with tele-dentistry platforms could further broaden access to specialized care in underserved areas, collectively underscoring the potential of AI to advance precision dentistry.

\begin{figure}
\centering
\includegraphics[width=1.0\linewidth]{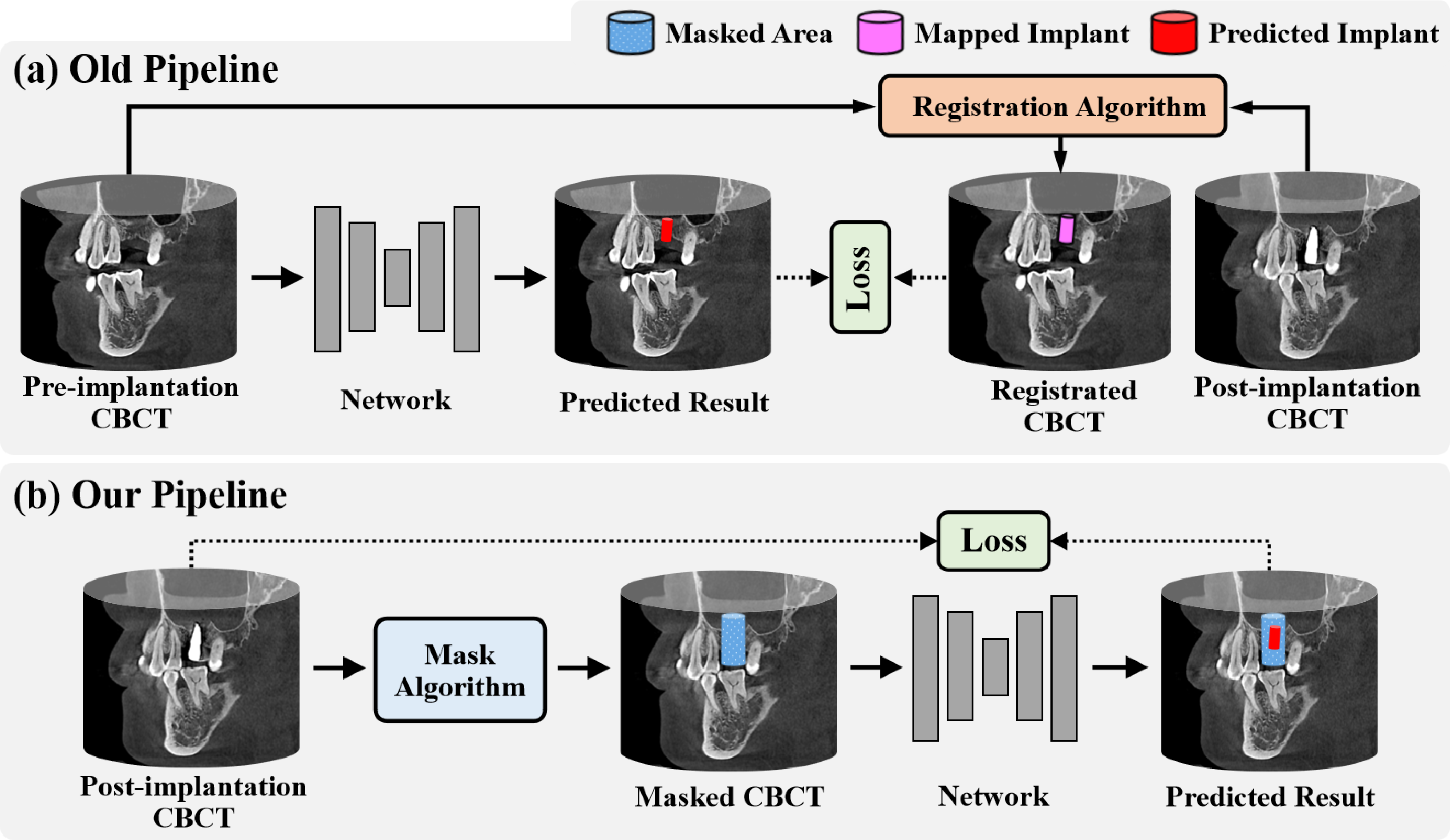}
\caption{Comparison of registration-based method and our registration-free method. (a) CBCT data. (b) Previous implant position extraction. (c) Our registration-free implant position extraction. 
} \label{fig_illusion}
\end{figure}

Current learning-based implant planning methods employ registration-based pipelines to address the absence of ground-truth labels in pre-implantation scans and metallic artifacts in post-implantation scans. These methods align post- with pre-implantation data to generate training labels (Fig.~\ref{fig_illusion}(a)). Approaches are categorized by input strategy: 2D methods utilize panoramic radiographs~\cite{mun2025ai} or CBCT slices~\cite{widiasri2022dental,yang2022implantformer} for segmentation and regression tasks, offering computational efficiency at the cost of 3D context; 3D methods process full CBCT volumes as video sequences~\cite{yang2024simplify} or via 3D segmentation~\cite{cai2025intelligent} to preserve spatial features.

Despite the progress, the prevailing registration-based approach faces two primary challenges: 1) registration accuracy is sensitive to anatomical changes (\eg, tooth extraction); and 2) the scarcity of paired pre- and postoperative CBCT scans, which constrains the construction of large-scale implant planning datasets.
To tackle these, we introduce a \textbf{Reg}istration-\textbf{Free} \textbf{Net}work (\textbf{RegFreeNet}). The method builds on a critical clinical observation: implant positioning is largely determined by the morphology of adjacent teeth, which remains intact when the implant site is masked. Hence, we directly train on masked post‑implantation CBCT scans, eliminating the need for paired pre‑ and postoperative data (Fig.~\ref{fig_illusion}(b)). 
By directly leveraging masked postoperative scans as input, RegFreeNet offers two distinct advantages over conventional registration‑based approaches: 1) it avoids error‑prone registration entirely; and 2) it facilitates the assembly of large‑scale, multi‑source datasets without the need for costly data pairing.

Despite their advantages, RegFreeNet faces a critical challenge in mandibular implant planning: the masked scan obscures the mandibular nerve canal. Since safe placement requires avoiding this canal, the lack of paired pre-/postoperative data prevents the model from learning the detailed anatomical variations needed to infer the canal's position precisely. To tackle this, a \textbf{N}eighboring \textbf{D}istance \textbf{P}erception (\textbf{NDP}) module is proposed to implicitly learn anatomical hierarchies from postoperative scans, compensating for the absence of paired data. Specifically, it integrates multi-scale convolutional kernels to capture progressive structural variations across tooth regions and a subsequent graph convolution network to explicitly model spatial relationships among neighboring teeth. This dual mechanism enables adaptive encoding of both local tooth topology and global jaw morphology, reconstructing the structural context necessary for accurate mandibular canal localization. By restoring critical spatial supervision, the NDP module provides reliable anatomical grounding for implant planning, allowing registration-free frameworks to achieve nerve canal localization precision comparable to registration-based approaches while maintaining clinical safety constraints.

The second challenge for RegFreeNet is the absence of explicit angular supervision for implant inclination, which introduces geometric ambiguity. This under-constrained formulation allows positional errors to propagate into clinically unacceptable inclination deviations, jeopardizing prosthetic outcomes. 
To address this, we propose a dual-branch architecture that decomposes implant placement into two explicitly regularized sub-tasks: a position-regression branch and a slope-prediction branch. The slope branch is trained directly on inclination vectors from post-implantation CBCT and serves as an explicit geometric regularizer to constrain the position branch, thereby injecting anatomical priors and resolving the inherent geometric ambiguity of unpaired learning. 
Unlike previous registration-free methods that entangle position and orientation predictions, our approach explicitly disentangles and supervises both outputs through separate branches. This design not only stabilizes training through well-constrained sub-problems but also provides explicit geometric regularization and enhances clinical interpretability, yielding more anatomically plausible and reliable implant plans than end-to-end unpaired approaches.

Our main contributions are four-fold:
1)~We propose a novel registration-free framework that operates directly on masked post-implantation CBCT volumes, eliminating alignment errors inherent in registration-based methods. 
2)~Within this framework, we present a dual-branch architecture that explicitly decouples and independently supervises implant position and inclination, removing dependence on paired preoperative data while enhancing prediction interpretability.
3)~To resolve structural ambiguity in registration-free learning, we design a Neighboring Distance Perception module that captures anatomical context through multi-scale feature extraction.
4)~To support evaluation, we introduce and release ImplantFairy, the first public large-scale dental implant dataset with 1,622 CBCT scans. Extensive experiments across three benchmark datasets demonstrate that our method outperforms nine state-of-the-art approaches in positioning accuracy.

\section{Related Work}
\subsection{Dental Implant Planning}

Research in computer-aided implant planning has advanced through three methodological generations, each progressively addressing clinical challenges with more sophisticated computational techniques. 

Early expert systems establish the foundation of digital implantology by developing specialized software for surgical simulation. For example, Spector \etal created a planning platform that reconstructed 3D jawbone models from CBCT data, enabling virtual implant placement in relation to prosthetic requirements~\cite{spector2008computer}. Galanis \etal introduced optimization criteria to automatically determine implant size and position during preoperative planning~\cite{galanis2007computer}. Other systems adopt distinct strategies, such as Szejka \etal's interactive reasoning system for optimal implant length selection~\cite{szejka2011reasoning}, and Sadighpour \etal's artificial neural network for prosthesis-type decision support~\cite{sadighpour2014application}. While these systems demonstrate the viability of computational assistance, they require substantial manual intervention and parameter tuning.

Deep learning-based 2D approaches emerge to address specific subtasks in implant planning using convolutional architectures. For example, Kurt \etal employed multi-stage CNNs for tooth segmentation and virtual mask generation based on adjacent tooth anatomy~\cite{kurt2021deep}. Dental-YOLO~\cite{widiasri2022dental} detects critical anatomical structures in sagittal CBCT slices to assess bone dimensions. Subsequent works incorporate multimodal data, such as Mun \etal’s combination of panoramic radiographs with clinical information for implant quantity prediction~\cite{mun2025ai}. Other recent advances include transformer-based regression networks that integrates CLIP-based textual guidance~\cite{yang2022implantformer,yang2023tceip,yang2023tcslot}. However, these methods’ reliance on 2D representations inherently limits their spatial reasoning capabilities and full 3D contextual awareness. 

Volumetric processing methods were subsequently developed to overcome the limitations of 2D representations by preserving 3D contextual information. For instance, Yang \etal reformulated CBCT sequence analysis as a video processing task to maintain spatiotemporal feature coherence across adjacent slices~\cite{yang2024simplify}. Segmentation-driven approaches, such as those by Al-Asali \etal~\cite{al2024deep} and Cai \etal~\cite{cai2025intelligent}, utilize nnU-Net architectures to infer implant positions through shape completion followed by mathematical optimization. While improving 3D anatomical awareness, these methods remain fundamentally dependent on registration-based frameworks, which require precisely aligned pre- and postoperative scans and are susceptible to cumulative alignment errors.

Indeed, a fundamental limitation across existing methods is their dependence on registered scan pairs, which hinders scalability and practical deployment due to the stringent requirement for precisely aligned data. In contrast, we propose a registration-free paradigm that eliminates this constraint, facilitating large-scale training while directly addressing the challenges of inclination estimation and anatomical context integration through a dedicated network architecture.

\subsection{CBCT Datasets}
Cone-beam computed tomography (CBCT) has become an indispensable imaging modality for 3D maxillofacial analysis. While the field has seen substantial advances in AI algorithm development, current research priorities exhibit a significant imbalance: considerable effort is directed toward achieving high performance on curated datasets, while fundamental questions regarding dataset suitability and scalability remain inadequately addressed, as noted by Sengupta \etal~\cite{sengupta2022scarcity}.

As compared in Tab.~\ref{table_dataset}, existing public CBCT repositories are predominantly designed for anatomical segmentation. The foundational work by Cipriano \etal~\cite{cipriano2022deep} introduces the first dedicated mandibular canal segmentation dataset, which later evolves into the ToothFairy~\cite{bolelli2024segmenting} and ToothFairy2~\cite{bolelli2025segmenting} benchmarks by Bolelli \etal. Similarly, the CTooth dataset by Cui \etal~\cite{cui2022ctooth,cui2022fully} focuses exclusively on tooth structure segmentation. However, these datasets present critical limitations for implant positioning research: they either provide incomplete volumetric information, such as CTooth's partial CBCT scans, or lack essential annotations required for spatial planning tasks. To support comparative external validation, we curate subsets containing implant placements from Cui \etal (61 scans) and Bolelli \etal (51 scans). Nevertheless, a fundamental mismatch persists: these resources were designed for segmentation objectives, not for the spatially grounded challenge of predicting implant position and inclination.

This underscores the distinct contribution of our ImplantFairy dataset, which provides comprehensive, implant-centric annotations specifically tailored for developing and evaluating  dental implant planning systems.

\begin{table}[!t]
\caption{Comparison of CBCT datasets. * denotes different versions of the same dataset.}
\centering
\begin{tabular}{l|c|l|r|r|r}
\toprule
Dataset      & Year & Country            & Train & Test & Public \\ \hline
Cipriano \etal\cite{cipriano2022deep} & 2022 & ITA              & 332                 & 15              &  347             \\ \hline
Cui \etal~\cite{cui2022fully}     
& 2022 & CHN              & 4,531                & 407            & 150               \\\hline
CTooth~\cite{cui2022ctooth}   
& 2022 & CHN              & 22                  & -               & 22             \\ \hline
ToothFairy~\cite{bolelli2024segmenting} 
& 2023 & ITA, NL & 443                 & 50            &  493             \\ \hline
ToothFairy2~\cite{bolelli2025segmenting} 
& 2024 & ITA, NL & 480                 & 50              & 530              \\ \hline
ImplantFairy (Ours)            & 2025 & CHN              & 1,369                    & 253     & 1,622                    \\ 
\bottomrule
\end{tabular}
\label{table_dataset}
\end{table}

\section{Proposed RegFreeNet}
\begin{figure*}
	\centering
	\includegraphics[width=1.0\linewidth]{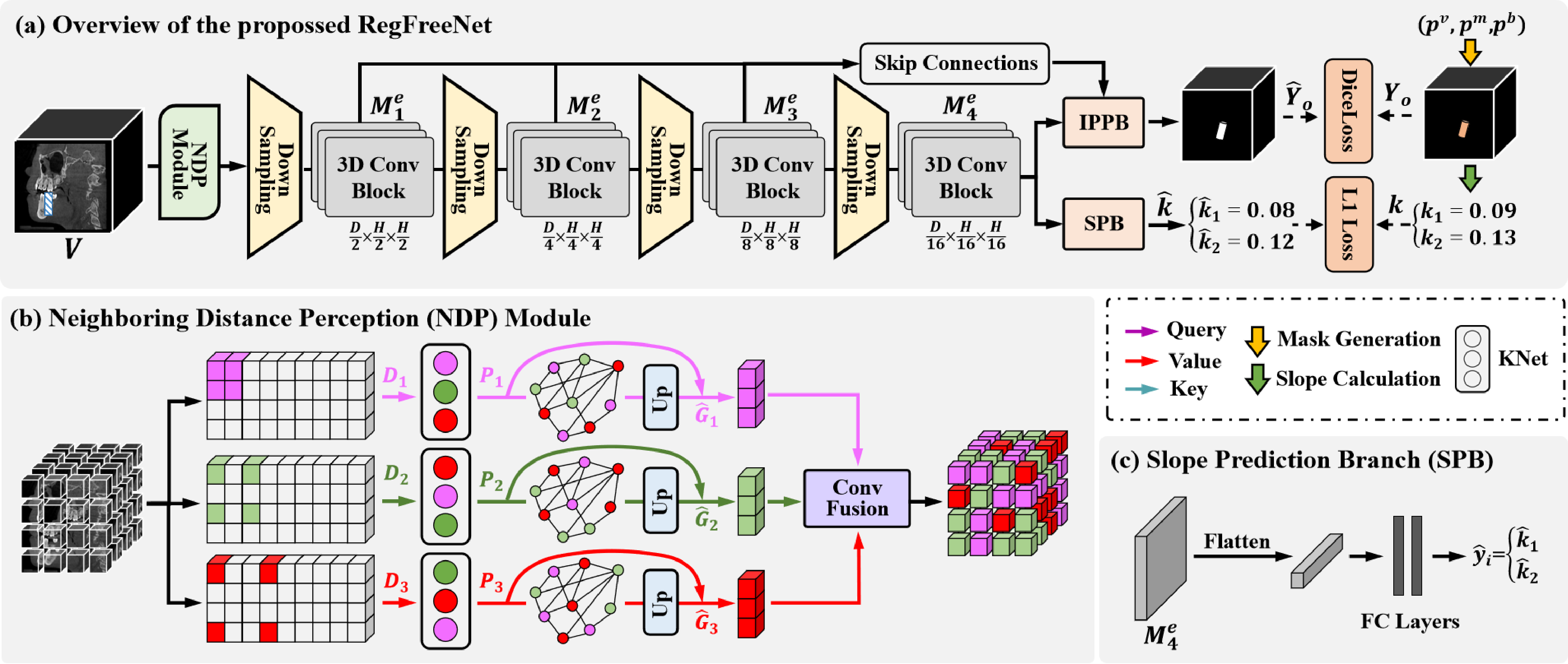}
	\caption{(a) Overview of the propossed RegFreeNet, which takes the CBCT volume as input and predicts the implant position and slope simultaneously. (b) Neighboring Distance Perception (NDP) Module, which captures fine-grained structural variations using multi-scale convolutional kernels, followed by a graph convolution network to explicitly model spatial relationships among neighboring teeth. (c) Slope Prediction Branch. 
    } \label{fig_network}
\end{figure*}
\subsection{Overview of Proposed RegFreeNet}

Conventional registration-based approaches for implant planning are constrained by several fundamental limitations that impede their clinical applicability. First, the accuracy of image registration is highly sensitive to anatomical changes (\eg, tooth extraction or occlusal wear) often resulting in misalignment between pre-operative and post-operative scans. Second, the prevalent clinical practice of utilizing panoramic radiographs rather than post-operative CBCT leads to incomplete data pairs, complicating the construction of large-scale training datasets. Most critically, registration errors propagate through the geometric interdependence between implant position and inclination, potentially yielding clinically significant deviations that compromise prosthetic functionality.

To overcome the limitations of registration-based methods, we introduce a \textbf{Reg}istration-\textbf{Free} \textbf{Net}work (\textbf{RegFreeNet}) based on a novel implant masking strategy. This approach occludes the implant region in post-implantation CBCT scans, creating a self-supervised learning paradigm that eliminates the need for paired pre- and post-operative data. However, the masking process inherently removes explicit spatial and contextual supervision, leading to two critical challenges: the loss of anatomical context around the implant site and the difficulty in modeling the geometric relationship between implant position and inclination without ground-truth references. To address these, our proposed RegFreeNet (Fig.~\ref{fig_network}) incorporates two key innovations. First, a \textbf{N}eighboring \textbf{D}istance \textbf{P}erception (\textbf{NDP}) module is integrated at the encoder input to compensate for missing anatomical context. It captures fine-grained structural variations (\eg, inter-tooth distance progression from crown to root) using multi-scale convolutional kernels to model progressive regional changes, followed by a graph convolution network that explicitly encodes spatial relationships among neighboring teeth. By jointly learning local tooth topology and global jaw morphology, the NDP module reconstructs the anatomical context necessary for accurate mandibular canal localization in registration-free settings.  
Second, the decoder adopts a dual-branch architecture. In addition to the position regression branch, a dedicated Slope Prediction Branch operates in parallel to explicitly model the geometric interdependence between implant position and clinical inclination. This design transforms an otherwise under-constrained prediction task into a well-regularized optimization problem, where slope prediction serves as auxiliary supervision to enforce anatomical plausibility. Together, these components enable accurate and clinically viable implant predictions from single postoperative scans.

The proposed RegFreeNet presents three significant advantages over registration-based methods. First, by eliminating the dependency on registration, it facilitates large-scale training with diverse, unpaired CBCT scans from multiple clinical centers, thereby enhancing model generalization. Second, the slope-aware learning mechanism explicitly models the geometric relationship between position and inclination, transforming an under-constrained problem into a well-regularized optimization that mitigates clinical risk. Third, its multi-scale perceptual capacity adapts to anatomical variations across patients and dental configurations. Together, these contributions establish a more scalable, robust, and clinically viable paradigm for implant positioning, advancing beyond the constraints of conventional registration-dependent approaches.

\subsection{Neighboring Distance Perception Module}
The registration-free framework removes explicit spatial supervision, posing a critical challenge: conventional 3D encoders with uniform convolutional kernels cannot adequately capture the hierarchical anatomical structure needed for implant positioning. In particular, they fail to perceive the progressive increase in inter-tooth distance from crown to root, a key spatial relationship guiding implant placement. To address this limitation, the proposed NDP module incorporates two complementary mechanisms designed to capture fine-grained local anatomical details and coherent global structural relationships. 

\subsubsection{Multi-Scale Dilated Convolutions}
Specifically, the NDP module replaces the initial encoder layer with a multi-scale feature extraction architecture, as illustrated in Fig.~\ref{fig_network}, three parallel dilated convolutions~\cite{dai2017deformable} with rates $n=(2,3,4)$ are applied to capture features across complementary receptive fields: 
\begin{equation}
\bm{D}_j=\textsc{Conv}_j(\bm{V}), \quad j\in(1,2,3)
\end{equation}
where $\bm{V} \in \mathbb{R}^{C\times D\times H\times W}$ denotes the input CBCT volume. 
We conducted an ablation experiment to determine a suitable $n$, details are given in Section V.

This design enables simultaneous perception of fine-grained local topology and coarse-grained global spatial trends. 
However, the learned features from multi-scale dilated convolutions remain in the image domain, which inherently limits their ability to explicitly model the complex spatial relationships between anatomical structures. 
This limitation motivates the introduction of a keypoint network (KNet) and a graph convolution network (GCN) to transform image features into a more structured representation.

\subsubsection{Graph Convolution Network} 
Then, the features $\bm{D}_j$ are fed into a keypoint network (KNet) to transfer the image feature to point feature $\bm{P}$:
\begin{equation}
\bm{P}_j=\textbf{KNet}(\bm{D}_j), j\in(1,2,3)
\end{equation}

Specifically, KNet employs a 3D convolution and an adaptive average pooling to extract a fixed set of 64 nodes (4×4×4) from $\bm{D}_j$, which effectively reduces the spatial complexity while preserving the most salient anatomical information. This transformation enables the subsequent graph convolution networks (GCN) to operate on a graph structure where nodes represent key anatomical locations and edges encode their spatial relationships.
Then, we employ three independent GCNs to take $\bm{P}_j$ as input to learn the spatial relationship between neighboring teeth:
\begin{equation}
\bm{G}_j=\textbf{GCN}(\bm{P}_j), j\in(1,2,3)
\end{equation}

GCN consists of two fully connected layers, which performs message passing between different nodes, allowing the model to learn explicit spatial constraints and anatomical priors. 
This graph-based representation provides a more natural and interpretable framework for modeling the hierarchical anatomical structure of the dentition, overcoming the limitations of purely convolutional approaches in capturing complex spatial dependencies.
After the spatial location learning in GCN, we transform $\bm{G}_j$ into image like features $\bm{\hat G}_j$ by upsample operation.
To constrain the effects of graph convolution transformations, we fuse the features of the original points with $\bm{\hat G}_j$ via residual connection.

The fused features are integrated via a \(1\times1\times1\) convolution, producing a comprehensive representation that encodes spatial relationships critical for implant positioning. 
By explicitly modeling anatomical constraints, the NDP module enables more plausible predictions and enhances generalization across diverse dental anatomies.

\subsubsection{3D U-Net Encoder} 
Subsequent feature encoding employs a conventional 3D U-Net backbone, which has proven effectiveness in medical image segmentation tasks, particularly for volumetric data like CBCT scans.
It's structured as four sequential encoding blocks. Each block implements two convolutional operations followed by max pooling, systematically reducing spatial dimensionality through four downsampling stages. 
This configuration progressively reduces spatial resolution while increasing feature channel depth, enabling the network to capture both fine-grained local details and coarse-grained global context. 
The four encoding stages produce feature maps at resolutions of 1/2, 1/4, 1/8, and 1/16 of the original input size, with corresponding channel dimensions of 64, 128, 256, and 512, respectively. 
These hierarchical representations $\{\bm{M}^e_1, \bm{M}^e_2, \bm{M}^e_3, \bm{M}^e_4\}$ encode multi-scale anatomical information essential for understanding the complex spatial relationships in dental CBCT volumes.

\subsection{Dual-Task Regression} 
The registration-free paradigm faces a fundamental challenge: the lack of explicit spatial correspondence eliminates direct supervision for modeling two clinically critical and interdependent parameters: implant position and inclination. Notably, inclination can often be estimated more reliably from anatomical landmarks, even when precise positional cues are absent. By leveraging this inherent asymmetry, wherein inclination provides a geometrically stable reference, we regularize the more ambiguous position prediction through a dual-branch framework. Our approach explicitly models the coupling between position and slope via parallel, interconnected regression branches, transforming an ill-posed problem into a well-constrained optimization that mitigates error propagation and enhances anatomical plausibility.

\begin{figure*}
\centering
\includegraphics[width=1\linewidth]{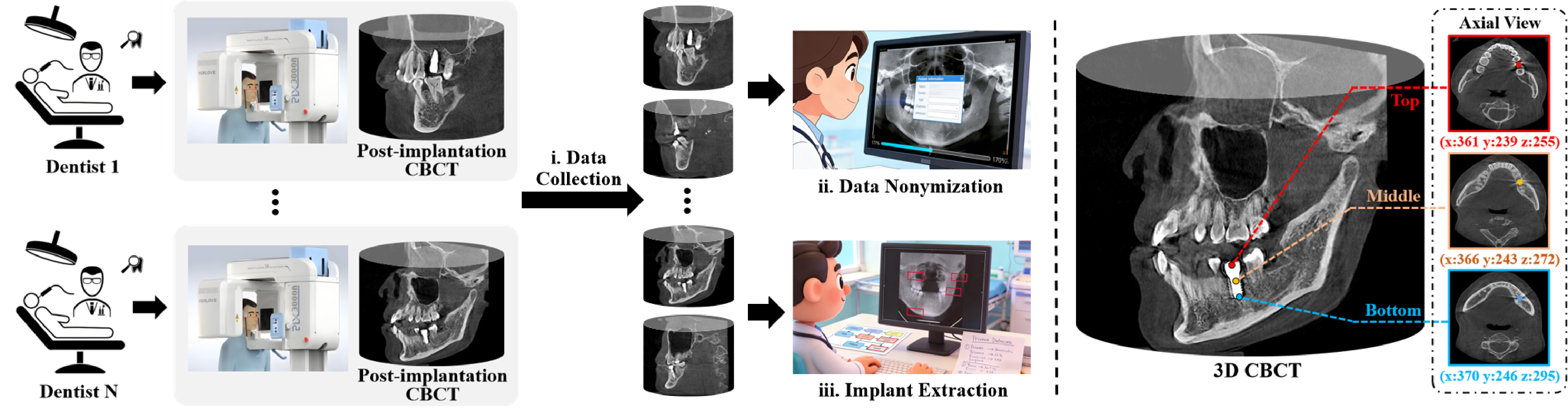}
\caption{\textbf{Left:} The dataset pipeline of ImplantFairy. \textbf{Right:} The determination process of implant position. The red, orange, and blue dashed lines represent the apex, midpoint, and bottom of the implant, respectively.} \label{fig_dataset_process}
\end{figure*}

\subsubsection{Implant Position Prediction Branch}
Accurate implant localization presents significant challenges owing to the volumetric complexity of CBCT data and inherent anatomical variability among patients. 
Whereas conventional methods depend on single-point regression or single-slice segmentation-based localization, they frequently exhibit limited spatial consistency across sectional views. 
Therefore, we modulate the task of implant position prediction as 3D volume segmentation, which not only improves localization precision but also encodes essential structural information for inclination estimation. 
Moreover, this paradigm substantially reduces cumulative errors relative to slice-wise prediction strategies, thereby yielding clinically viable implant trajectories.
Specifically, we generate the corresponding 3D implant label based on the implant label $\bm{p}$.
First, we generate an all-zero matrix.
Starting from the implant's vertex \(\bm{p}^v\), we generate a circle with a radius R on each slice of CBCT, extending to the implant's bottom point \(\bm{p}^b\).
In our experiment, R is set to 14, which is the average value of clinical implants. Pixels within the circle are assigned a value of 1.

We design an implant position prediction branch (IPPB) to locate the implant. The architectural implementation comprises four upsampling convolution stages followed by a prediction head. Starting from the lowest-resolution encoder feature map $\bm{M}^e_4$, the network performs iterative upsampling and fusion with corresponding encoder features $\bm{M}^e_3$ through $\bm{M}^e_1$, progressively restoring spatial resolution to match the input sub-volume dimensions. The prediction head then reduces channel depth to one, generating the final implant position output. Training utilizes a composite objective function combining cross-entropy and soft Dice loss:
\begin{align}
    \mathcal{L}_{\textsc{Seg}} 
	&= \mathcal{L}_{\text{Dice}} + \mathcal{L}_{\text{CE}} \\
	&=\left(1-\sum\frac{2\sum_{o=1}^{O} \bm{Y}_{o}\cdot\hat{\bm{Y}}_{o}}{\sum_{o=1}^{O} \bm{Y}_{o}^{2}+\sum_{o=1}^{O}\hat{\bm{Y}}_{o}^{2}}\right)-\sum_{o=1}^{O} \bm{Y}_{o}\log\hat{\bm{Y}}_{o}, 
\end{align}
where $O$ represent the number of voxels; $\bm{Y}_{o}$ and $\hat{\bm{Y}}_{o}$ denote the ground truth map and predicted probability map, respectively, at voxel $o$. The cross-entropy loss ensures precise voxel-wise classification accuracy by minimizing the divergence between predicted and ground-truth probability distributions, while the soft Dice loss directly optimizes spatial overlap between predicted and actual implant regions to enhance segmentation quality.

\subsubsection{Slope Prediction Branch}
The slope prediction branch (SPB) is designed to leverage anatomical constraints inherent in CBCT volumetric data. Clinically, implant inclination is determined by alveolar bone morphology and spatial relationships with adjacent dentition, but learning this correlation without paired pre-operative scans poses a fundamental challenge. Our approach formalizes slope regression as an auxiliary task, analytically deriving inclination parameters \(k^1\) and \(k^2\) from the implant’s 3D trajectory via a spatial transformation algorithm. While this branch shares the feature input \(\bm{M}^e_4\) with the position regression module, it employs dedicated fully connected layers to decode angular information. The core contribution is the use of slope prediction as an implicit regularizer,  enforcing geometric consistency between position and orientation outputs guides the network toward anatomically grounded representations, compensating for the lack of explicit spatial supervision in registration-free learning and improving robustness while reducing prediction outliers.

\begin{figure*}
\centering
\includegraphics[width=1.0\linewidth]{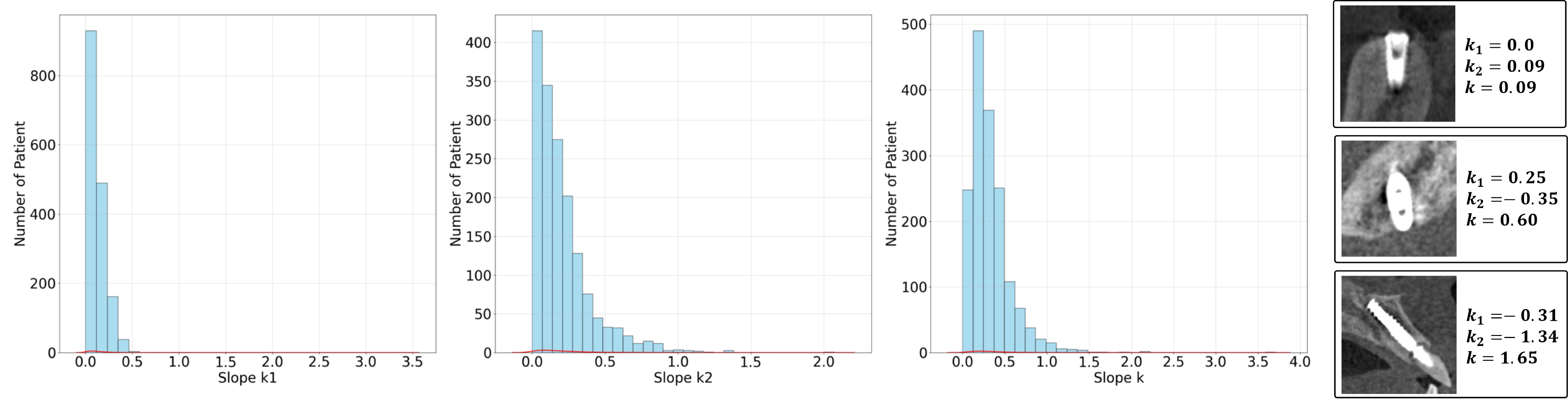}
\caption{The slope distribution in ImplantFairy.} \label{fig_slope}
\end{figure*}

Specifically, given \(N_j\) 3D coordinates \((x_i, y_i, z_i)\) for \(i=1,2,\dots,N_j\) of an implant, the ground-truth slope parameters \(\bm{k}_j = [k_j^1, k_j^2] \) can be easily computed following the spatial transformation formulation established in Implantformer~\cite{yang2022implantformer}:
\begin{align}
	k_j^1=\frac{N_j \sum_{i=1}^{N_j} x_i z_i-\sum_{i=1}^{N_j} x_i \times \sum_{i=1}^{N_j} z_i}{N_j \sum_{i=1}^{N_j} z_i^2-\sum_{i=1}^{N_j} z_i \times \sum_{i=1}^{N_j} z_i}, \\
	k_j^2=\frac{N_j \sum_{i=1}^{N_j} y_i z_i-\sum_{i=1}^{N_j} y_i \times \sum_{i=1}^{N_j} z_i}{N_j \sum_{i=1}^{N_j} z_i^2-\sum_{i=1}^{N_j} z_i \times \sum_{i=1}^{N_j} z_i},
\end{align} 
The slope distribution across the ImplantFairy dataset (Fig.~\ref{fig_slope}) reflects clinical norms, with values predominantly falling within 0–0.4 radians, consistent with biomechanical stability requirements—a finding that substantiates the physiological relevance of our regression target.

As illustrated in Fig.~\ref{fig_network}, the slope branch first flattens the encoded feature map $\bm{M}^e_4$ into a vector, which is subsequently processed through two fully-connected (FC) layers to regress the slope parameters:
\begin{equation}
	\bm{\hat k} = \textbf{FC}(\textbf{flatten}(\bm{M}^e_4))
\end{equation}
where the first FC layer projects the flattened feature vector to a 256-dimensional hidden representation, followed by ReLU activation and dropout with probability 0.5 to prevent overfitting. The second FC layer then maps this representation to the final 2-dimensional output $\bm{\hat k}=[\bm{\hat k^1}, \bm{\hat k^2}]$, representing the predicted slope parameters in the x-z and y-z planes, respectively.

The optimization objective is defined using \(L_1\) loss: 
\begin{equation}
	\mathcal{L}_\textsc{Slope} = \sum_j\|\bm{k}_j - \hat{\bm{k}}_j\|,
\end{equation}
where \(\bm{k}_j\) and \(\hat{\bm{k}}_j\) denote the ground-truth and predicted slope values, respectively. 
The slope prediction branch significantly reduces position prediction errors, particularly in cases with complex anatomical variations, by providing additional geometric constraints that guide the network toward clinically plausible solutions.

The final training loss of RegFreeNet is given as:
\begin{equation}
	\mathcal{L} = \mathcal{L}_\textsc{Seg}+\mathcal{L}_\textsc{Slope},
\end{equation}

\section{Construction of ImplantFairy Dataset}
\subsection{Overview of Construction Pipeline}
The absence of large-scale, publicly available datasets has significantly impeded the advancement of AI-driven dental implant planning, primarily due to the reliance on complex and often infeasible registration protocols. By leveraging our proposed registration-free framework (RegFreeNet), we introduce ImplantFairy—the first comprehensive dental implant dataset constructed without dependency on pre- and post-operative scan alignment. As shown in Fig.~\ref{fig_dataset_process}, the dataset curation process encompasses data acquisition, anonymization, annotation generation, and rigorous quality control, establishing a new benchmark for implant positioning research.

\subsection{Dataset Acquisition}
ImplantFairy comprises 1,622 CBCT scans collected from the Department of Stomatology at Shenzhen University General Hospital (SUGH), a tertiary care institution specializing in oral implantology and maxillofacial surgery. All scans were acquired using a KaVo 3D eXami system (Imagine Sciences International LLC) under a standardized imaging protocol: two sequential 20-second scans per patient, yielding volumetric data with an isotropic voxel resolution of 0.2 mm. The dataset is partitioned into 1,369 scans for model training and 253 scans for robustness evaluation, ensuring sufficient data diversity and clinical representativeness.

The compilation and use of the ImplantFairy dataset were conducted in accordance with ethical guidelines and approved by the Institutional Review Board of Shenzhen University General Hospital. All sensitive personal identifiers were removed prior to dataset assembly, retaining only non-identifiable imaging data and basic demographic variables (gender and age) for research purposes.

\subsection{Annotation Generation}
Annotation was performed using a streamlined three-point landmarking approach. For each implant, clinicians identified three key coordinates: the vertex \(\bm{p}^v\), midpoint \(\bm{p}^m\), and base point \(\bm{p}^b\) (Fig.~\ref{fig_dataset_process}) along the implant’s longitudinal axis. These annotations were subsequently converted into 3D volumetric labels by generating a series of cylindrical masks aligned with the implant trajectory. Starting from \(\bm{p}^v\) and extending to \(\bm{p}^b\), a circle of radius \(R=14\) voxels (reflecting the average implant diameter in the dataset) was projected axially across successive slices. Voxels within these circular regions were assigned a value of 1, producing a continuous 3D implant representation suitable for volumetric learning. 

\subsection{Data Statistics}
The dataset includes partially anonymized metadata, preserving only gender and age information. Among the included subjects, 59.20\% are male, with a gender distribution of 57.20\% in the training set and 42.80\% in the test set. All scans were obtained between 2021 and 2025. Patient ages range from 10 to 100 years, with peak frequencies in the 20–30 and 60–70 age groups in the training set, and 50–70 in the test set, reflecting typical dental implant demographics.

\subsection{Evaluation Protocols}
We employ two evaluation metrics, i.e., the Dice similarity coefficient (Dice) and Intersection over Union (IoU) to evaluate the performance of the different methods. 
The Dice coefficient quantifies the overlap between the volumetric segmentation predictions $\hat{\mY}$ and the corresponding ground truth map $\mY$, which is expressed as follows:
\begin{equation}
	\text{Dice}(\hat{\mY}, \mY) = \frac{2|\hat{\mY} \cap \mY|}{|\hat{\mY}| + |\mY|}
\end{equation}
It ranges from 0 to 1, where a value of 1 indicates perfect overlap between the predicted segmentation and the ground truth. 
This coefficient provides a measure of the overlap between the predicted and true regions.

Similarly, IoU measures the overlap by computing the ratio of the intersection area to the union area of the prediction and the ground truth:
\begin{equation}
	\text{IoU}(\hat{\bm{Y}}, \bm{Y}) = \frac{|\hat{\bm{Y}} \cap \bm{Y}|}{|\hat{\bm{Y}} \cup \bm{Y}|}
\end{equation}
The IoU score also ranges from 0 to 1, with 1 representing a perfect match. These two metrics are commonly used together to provide a comprehensive evaluation of prediction accuracy.

\section{Experimental Results}
\subsection{Experimental Setup}
\subsubsection{Compared Methods}
We compare our RegFreeNet against the state-of-the-art segmentation methods, covering CNN-based and Transformer-based architectures. 
The compared CNN-based methods include 3DUNet~\cite{ronneberger2015u}, 3DUNet++~\cite{zhou2018unet++}, DAF3D~\cite{wang2019deep}, VNet~\cite{milletari2016v} and UXNet~\cite{lee20223d}. 
The Transformer-based methods include UNETR~\cite{hatamizadeh2022unetr}, UNETR++~\cite{shaker2024unetr++}, SwinUNETR~\cite{hatamizadeh2021swin}, and SwinUNETRv2~\cite{he2023swinunetr}. 
The transformer-based methods utilize Vision Transformer~\cite{dosovitskiy2020image} or SwinTransformer~\cite{liu2021swin} as encoders to learn global features for prediction.
To ensure fairness, we use publicly available codes of all these methods and train them under the same training settings.

\subsubsection{Datasets}
Besides our in-house dataset, we collected CBCT scans containing implants from two publicly available datasets to further validate the model's performance.
Specifically, we extracted 12 CBCT scans from the dataset proposed by Cui \etal~\cite{cui2022fully} and ToothFairy2~\cite{bolelli2025segmenting} to construct an external implant dataset.
These CBCT scans were performed the same masking algorithm like test set.
To verify the robustness of the model, we retain the original scale of these data and do not perform a uniform scale transformation.
The external dataset is used to verify the generalization and robustness of the model, so it does not participate in the training of the model and is only used for testing.

\subsubsection{Implementation Details}
Pytorch is used for model training and testing. 
All model training and inference are operated on the platform of NVIDIA A40 GPUs. 
To prevent information leakage and enhance robustness, we introduce a random masking algorithm that applies a spatial offset to the implant mask during training, forcing the network to learn from variably occluded contexts. 
For the network training, we use a batch size of 4. 
The initial learning rate is set as 3e-4, with a linear warmup, cosine annealing learning rate scheduler, and AdamW optimizer with 5e-5 weight decay. 
Due to the particularity of the oral structure, it does not have the ability to rotate without deformation.
Therefore, we do not use additional data augmentation strategy, only the random crop is applied.
Specifically, input data are randomly cropped out at sizes of 128 × 128 × 128.
In our experiments, random cropping is performed after data splitting. Therefore, image slices originating from the same patient scan are kept exclusively within either the training set or the testing set, avoiding data leakage.
At inference time, we utilize test-time augmentation (TTA) methods~\cite{shanmugam2021better} (i.e., overlapped sliding window inference~\cite{isensee2021nnu}), with an overlap ratio set to 0.25.
MONAI library and ImageNet pre-training models are used in model training and inference.

\subsection{Comparison to the State-of-the-art Methods}
As shown in Table~\ref{table_mainstream}, the proposed RegFreeNet achieved optimal performance on the SUGH dataset, achieving a Dice score of 47.51 and IoU score of 0.3540, outperforming all other compared models. 
RegFreeNet also achieved the best performance on external datasets, achieving a Dice score of 33.0 and IoU of 0.2413. 
Notably, RegFreeNet demonstrates excellent generalization capabilities, performing significantly better than other models on external datasets.
Overall, Unet3D++ performed better among CNN-based methods, while SwinUNETR performed relatively well among Transformer-based methods, but their performance was lower than that of the proposed RegFreeNet. These experimental results validate the effectiveness of the RegFreeNet architecture.

\begin{table}[]
\caption{Ablation study of the NDP module and the slope prediction branch. The red and blue metrical scores denote the best and second-best quantitative perommance.}\label{table_ablation}
\centering
\begin{tabular}{c|cc|cc|cc}
		\toprule
		Method                 & \multicolumn{2}{c|}{Modules}   & \multicolumn{2}{c|}{SUGH}       & \multicolumn{2}{c}{External}   \\ \hline
		\multirow{5}{*}{RegFreeNet} & \multicolumn{1}{c|}{NDP} & SPB & \multicolumn{1}{c|}{Dice} & IoU & \multicolumn{1}{c|}{Dice} & IoU \\ \cline{2-7} 
		& \multicolumn{1}{c|}{\cxmark}    & \cxmark    & \multicolumn{1}{c|}{45.02}    & 0.3325    & \multicolumn{1}{c|}{22.10}     & 0.1404    \\ \cline{2-7} 
        
		& \multicolumn{1}{c|}{\ding{52}}    & \cxmark    & \multicolumn{1}{c|}{ 45.55}  & \color[HTML]{3531FF}0.3358    & \multicolumn{1}{c|}{\color[HTML]{3531FF}31.81}     & 0.1967    \\ \cline{2-7} 
        
		& \multicolumn{1}{c|}{\cxmark}    & \ding{52}    & \multicolumn{1}{c|}{\color[HTML]{3531FF}45.56}       & 0.3320    & \multicolumn{1}{c|}{ 31.64}     & \color[HTML]{3531FF} 0.1924    \\ \cline{2-7} 
        
		& \multicolumn{1}{c|}{\ding{52}}    & \ding{52}    & \multicolumn{1}{c|}{\color[HTML]{FE0000} 47.22}     & \color[HTML]{FE0000} 0.3555    &  \multicolumn{1}{c|}{\color[HTML]{FE0000} 31.87}     & \color[HTML]{FE0000} 0.2058    \\ 
		\bottomrule
\end{tabular}
\end{table}

\begin{table*}[]
	\caption{Comparison to the mainstream segmentation methods. The red and blue metrical scores denote the best and second-best quantitative perommance.}\label{table_mainstream}
	\centering
	\begin{tabular}{c|c|c|c|cc|cc}
		\toprule
		\multirow{3}{*}{Method}      & \multirow{3}{*}{Model} & \multirow{3}{*}{Param} & \multirow{3}{*}{FLops}& \multicolumn{2}{c|}{\multirow{2}{*}{SUGH}} & \multicolumn{2}{c}{\multirow{2}{*}{External Dataset}} \\
		&                        & \multicolumn{2}{c|}{}                      & \multicolumn{2}{c}{}                                  \\ \cline{5-8}  
		&     & &                    & \multicolumn{1}{c|}{Dice}      & IoU       & \multicolumn{1}{c|}{Dice}            & IoU             \\ \midrule
		\multirow{6}{*}{CNN-based}         & 3DUnet    &{\color[HTML]{FE0000}5.75} & 321.88             & \multicolumn{1}{c|}{\color[HTML]{3531FF} 45.02}     & \color[HTML]{3531FF}  0.3325    & \multicolumn{1}{c|}{22.10}           & 0.1404          \\ 
		& 3DUnet++    & 6.98 &  1335.7          & \multicolumn{1}{c|}{44.99}     & 0.3316    & \multicolumn{1}{c|}{22.31}           & 0.1338          \\ 
		& DAF3D      & 29.28 & 628.34            & \multicolumn{1}{c|}{43.43}     & 0.3023    & \multicolumn{1}{c|}{20.75}           & 0.1315          \\ 
		& VNet       & 45.60 &  759.58           & \multicolumn{1}{c|}{43.04}     & 0.3169    & \multicolumn{1}{c|}{\color[HTML]{3531FF} 26.24}           & \color[HTML]{3531FF} 0.1645          \\  
		& UXNet      & 53.01 &   1497.2          & \multicolumn{1}{c|}{41.45}     & 0.2934    & \multicolumn{1}{c|}{13.14}           & 0.0826          \\ 
		& RegFreeNet(our)    & {\color[HTML]{3531FF}5.82} &  269.25              & \multicolumn{1}{c|}{\color[HTML]{FE0000} 47.22}     & \color[HTML]{FE0000} 0.3555   & \multicolumn{1}{c|}{\color[HTML]{FE0000} 31.87}           & \color[HTML]{FE0000}{0.2058}         \\ \midrule
		\multirow{4}{*}{Transformer-based} & UNETR      &93.01 & \color[HTML]{3531FF}195.64             & \multicolumn{1}{c|}{41.76}     & 0.30    & \multicolumn{1}{c|}{20.06}           & 0.1198                 \\ 
		& UNETR++     & 122.1 & \color[HTML]{FE0000}147.04          & \multicolumn{1}{c|}{42.37}     & 0.3041    & \multicolumn{1}{c|}{21.42}                & 0.1285                \\ 
		& SwinUNETR   & 62.19 & 785.05           & \multicolumn{1}{c|}{44.18}     & 0.3188    & \multicolumn{1}{c|}{17.02}           & 0.1029          \\ 
		& SwinUNETRv2  & 72.76 & 846.27            & \multicolumn{1}{c|}{44.19}     & 0.3194    & \multicolumn{1}{c|}{20.69}           & 0.1297          \\ 
		\bottomrule
	\end{tabular}
\end{table*}

\subsection{Ablation Studies}
\subsubsection{Ablation of Key Components}
To validate the effectiveness of the proposed neighboring distance perception (NDP) module and slope prediction branch (SPB), we conducted an ablation study on both modules. Experimental results were given in Table.~\ref{table_ablation}. 
It can be observed from the table that the introduction of NDP and SPB can bring 1.87\% and 1.3\%  Dice score improvements on the SUGH dataset, and 1.33\%  and 0.27\%  improvements in IoU.
On the external dataset, the introduction of SPB brings significant performance improvements, with 8.52\%  Dice score improvement and 7.42\%  IoU improvement.
When NDP and SPB are integrated in RegFreeNet simultaneously, network performance is further improved, achieving the best performance.
The experimental results demonstrate the effectiveness of the proposed modules.

\section{Conclusion}
In this study, we develop a registration-free and slope-aware CBCT-based 3D implant position prediction network (RegFreeNet).
By masking the implants in the post-implantation data, we are able to discard the registration process and make it possible to construct a large-scale multi-center implant dataset.
On this basis, we propose ImplantFairy, a comprehensive, publicly accessible dental implant dataset with voxel-level 3D annotations of 1622 CBCT data. 
Due to the area variation characteristics of the tooth's spatial structure and the slope information of the implant, we designed a slope-aware implant position prediction network.
A neighboring distance perception (NDP) module is designed to adaptively extract tooth area variation features, and an implant slope prediction branch assists the network in learning more robust features through additional implant supervision information.
Extensive experiments conducted on ImplantFairy and the public dataset demonstrate that the proposed RegFreeNet achieves the state-of-the-art performance.

\bibliographystyle{IEEEtran}
\bibliography{ref}

\end{document}